\documentclass[conference]{IEEEtran}
\IEEEoverridecommandlockouts
\usepackage{cite}
\usepackage{amsmath,amssymb,amsfonts}
\usepackage{algorithmic}
\usepackage{graphicx}
\usepackage{textcomp}
\usepackage{xcolor}
\usepackage{hyperref}

\usepackage{booktabs}
\usepackage{adjustbox}
\usepackage{caption,subcaption}
\usepackage{breqn}
\newcommand\independent{\protect\mathpalette{\protect\independenT}{\perp}}
\def\independenT#1#2{\mathrel{\rlap{$#1#2$}\mkern2mu{#1#2}}}

\def\BibTeX{{\rm B\kern-.05em{\sc i\kern-.025em b}\kern-.08em
    T\kern-.1667em\lower.7ex\hbox{E}\kern-.125emX}}
\begin{document}

\title{Deep Personalized Glucose Level Forecasting Using Attention-based Recurrent Neural~Networks\\
}

\author{\IEEEauthorblockN{Mohammadreza Armandpour}
\IEEEauthorblockA{\textit{Department of Statistics} \\
\textit{Texas A\&M University}\\
College Station, USA \\
armand@stat.tamu.edu}
\and
\IEEEauthorblockN{Brian Kidd}
\IEEEauthorblockA{\textit{Department of Statistics} \\
\textit{Texas A\&M University}\\
College Station, USA \\
bkidd@stat.tamu.edu}
\and
\IEEEauthorblockN{Yu Du}
\IEEEauthorblockA{\textit{Biometrics Department} \\
\textit{Eli Lilly and Company}\\
Indianapolis, USA \\
du\_yu@lilly.com}
\and
\IEEEauthorblockN{Jianhua Z. Huang}
\IEEEauthorblockA{\textit{Department of Statistics} \\
\textit{Texas A\&M University}\\
College Station, USA \\
jianhua@stat.tamu.edu}
}

\maketitle

\begin{abstract}
 In this paper, we study the problem of blood glucose forecasting and provide a deep personalized solution. Predicting blood glucose level in people with diabetes has significant value because health complications of abnormal glucose level are serious, sometimes even leading to death. Therefore, having a model that can accurately and quickly warn patients of potential problems is essential. To develop a better deep model for blood glucose forecasting, we analyze the data and detect important patterns. These observations helped us to propose a method that has several key advantages over existing methods: 1- it learns a personalized model for each patient as well as a global model; 2- it uses an attention mechanism and extracted time features to better learn long-term dependencies in the data; 3- it introduces a new, robust training procedure for time series data. We empirically show the efficacy of our model on a real~dataset.
\end{abstract}
%We thoroughly analyze the data to find crucial patterns for improving our deep model.
\begin{IEEEkeywords}
Time series forecasting, Recurrent neural network, Personalized blood glucose prediction, Machine learning for diabetes
\end{IEEEkeywords}

\section{Introduction}
In 2017, more than 400 million people worldwide were living with diabetes. This number is expected to rise to 693 million by 2045 \cite{cho2018idf}. Predicting blood glucose (BG) levels would help diabetics both understand and control their BG levels to avoid complications \cite{atkinson2014type}. Machine learning algorithms have been widely applied to the prediction of glucose level with varying levels of success. Most of the early works provide prediction~based on a few observations per day \cite{albers2017personalized}. Recently, with the emergence of continuous glucose monotoring (CGM) devices, people can measure their blood glucose every 5 minutes, leading to much more data for models~to utilize.

The predictive models using only CGM data can be split into two prediction tasks: predicting risk of upcoming related health problems and forecasting BG levels. Some algorithms try to do hypo or hyperglycemia forecasting in the near future \cite{plis2014machine,oviedo2017review}. For predicting BG levels, the goal is not only to provide a final value prediction but also forecast trajectories. This allows patients to not only have an alert of possible BG events but also to understand the trend leading to it. Some trends require more immediate action to alleviate (e.g. the trend is continually accelerating downward). In this work, we will~focus on this more challenging and important task of~trajectory~prediction.

For these tasks, many models require added information (such as carbohydrate intake, insulin level, exercise, etc.) for prediction purposes \cite{de2012artificial,plis2014machine,turksoy2013hypoglycemia,zecchin2015jump,sudharsan2014hypoglycemia}. Although having more information can potentially improve the performance of the prediction, it necessitates that the patient collects the information. The data-collection process relies mainly on the subjective inputs provided by the person who wears a CGM device. This is an extra burden on the user and often results in inaccurate and noisy data. It has been shown that this data is often difficult to systematically collect and use in broader applications \cite{zecchin2016much,novara2015nonlinear}.
Hence, it is of significant importance to have a model that can both provide reliable prediction based solely on previous glucose levels and also take advantage of extra information whenever available. Our proposed method falls in this category. We train our model just based on the history of BG, as our application does not provide any exogenous variables. However, this information can be easily appended to the inputs of~our~model.

Various previous works have investigated methods for predicting BG levels. Some use classical approaches, like ARIMA~\cite{eren2010hypoglycemia,otoom2015real,boiroux2012overnight,novara2015nonlinear} and Random Forests \cite{sudharsan2014hypoglycemia,georga2015evaluation,hidalgo2017data}. Deep learning methods have shown competitive advantages to the classical models in many areas both for probabilistic~\cite{guen2020probabilistic,kurle2020deep} and deterministic~\cite{wu2020adversarial} prediction. Forecasting BG levels is no exception; two studies have shown the superior performance~\cite{fox2018deep,sun2018predicting} of deep learning. The downside is that they usually need lots of training examples to perform well. To overcome this, we develop a deep personalized forecasting model; it leverages data across different people to provide forecasting specific to each patient with fewer observations. This improves on the state of the art deep learning methods for BG levels \cite{fox2018deep,sun2018predicting}. They either ignore the individual variability in the forecasting or need lots of training examples per person to learn separate models for each patient.

After explaining our deep personalized model, we proceed by analyzing the data and showing the existence of long term dependencies within collected BG samples. The standard deep learning solution with recurrent neural networks (RNNs) cannot capture such a long dependency well because of the known problem of vanishing and exploding gradients \cite{pascanu2013difficulty}.
To overcome this issue, we add an \textit{attention mechanism} and time features to help the model capture this long term structure. Also, we observe high fluctuation in variables during training of the model, which is the result of high variance of batch gradients. To alleviate this issue, we exploit a more robust training procedure which happened to improve the final result as well. 
\begin{figure}
  \includegraphics[width=1\linewidth]{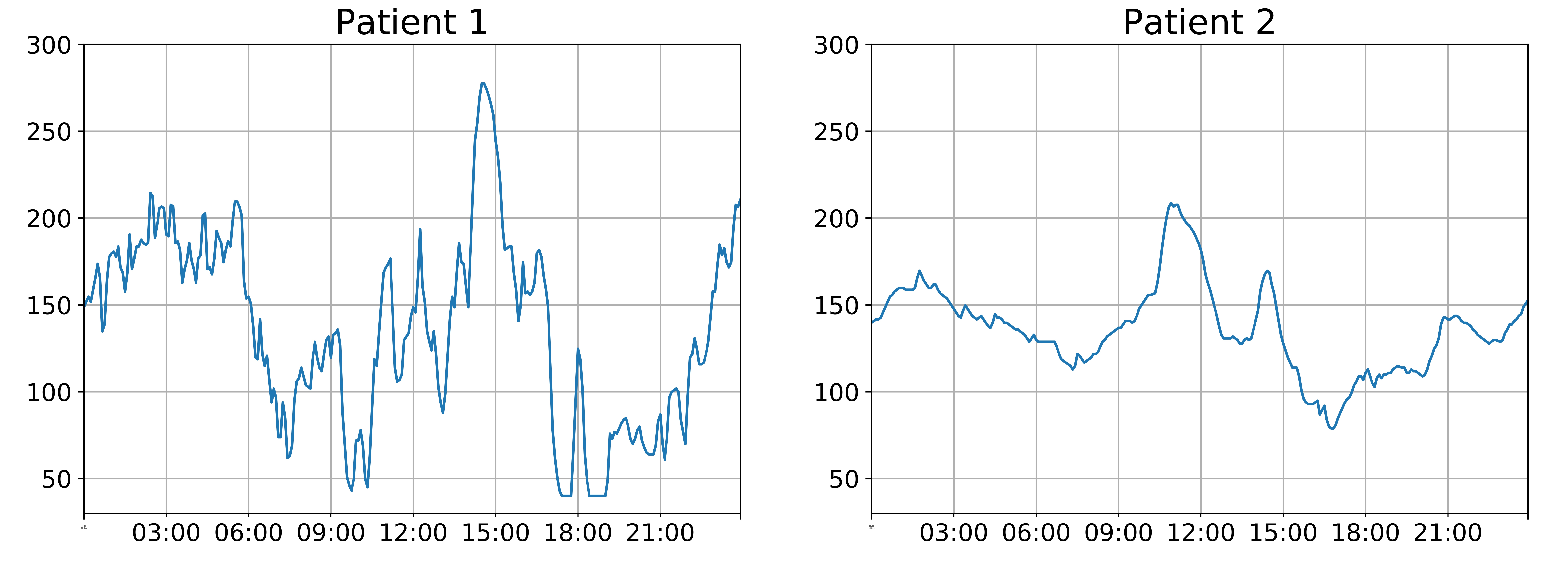}
  \caption{One day of CGM data for two different patients}
  %Two examples of CGM data for two patients over the course of 24 hours}
  \label{CGMex}
\end{figure}
Our main model is designed to provide BG forecasting, without the need for the user to manually enter any input, like carbohydrate intake or eating time. The model implicitly learns these events based on the previous behavior of each patient. We demonstrate the efficacy of our method over the existing state-of-the-art technique~\cite{fox2018deep} on 
a real dataset of BG values for different users provided by \cite{fox2018deep}.

\section{Methodology}

The goal of this section is to both provide some intuition behind our model and develop it in detail. We start by emphasizing the need for personalization, followed by a mathematical formulation of our personalized forecasting problem. Then we describe a deep learning component to capture the long term dependencies in our application and conclude the section with two further additions (time features and robust training).

\subsection{Personalization}

We start by motivating the importance of having a deep personalized model for the BG forecasting problem. Figure \ref{CGMex} shows the BG levels of two different diabetic patients over the course of one day. The plots illuminates the systematically different behavior of the two patients. The BG level of the second patient is clearly smoother and has less fluctuation than the first patient. This phenomena clearly suggests a single model for forecasting both patients is insufficient. On the other side, although deep learning based approaches have shown significant improvement over the traditional methods, in the existing literature they either ignore the individual identity and provide a generic model \cite{fox2018deep} or they learn an entirely new model for different patients \cite{zhu2018deep}. The latter is not an efficient solution because deep learning based models need lots of training examples; hence, we need many BG observations for each patient to provide useful predictions, which~is often not possible. In the following, we propose our deep personalized solution that alleviates the aforementioned problem by leveraging information across the different patients.

To accomplish the personalization while also allowing the model to share parameters across patients, we need to give the model information about which person is relevant for the specific BG measurements. The traditional way of giving the identity of the person to the model is by using a one-hot encoded vector. This one-hot encoded vector is large, sparse, and inefficient. Also, it does not represent any semantic information about the person. To overcome this inefficiency, we represent each person with a latent vector. The idea of using latent vectors, or embeddings, to represent identities in the model has shown great success in a wide range of topics, from social network analysis\cite{hoff2002latent,hamilton2017inductive,armandpour2019robust} to natural language processing \cite{mikolov2013distributed}. Some methods learn the embedding vectors manually using feature engineering, others use pre-trained embeddings from other tasks, and still others learn them by adding an embedding layer to the model, which we use in this paper. We will provide the details about our method and how we use the embedding layer in the following~section.

\subsection{Deep Personalized Forecasting}

In the following, we formulate learning a personalized forecasting model as a maximum likelihood optimization problem. Let $V$ to be the set of individuals and $g: V \to \mathcal{R}^d $ be the embedding layer from one-hot encoded vectors to latent feature representations with dimension $d$, which we aim to learn. Let $\{x^{(v)}_{1:T_v}\}_{v\in V}$ be a set of univariate time series, where $x^{(v)}_{1:T_v}=(x^{(v)}_{1}, x^{(v)}_{2}, \hdots, x^{(v)}_{T_v})$, and $x^{(v)}_{t} \in \mathcal{R}$ denotes the value of time series of person $v$ at time step $t$.

We proceed by defining our objective function based on the log-probability of observing $x^{(v)}_{T+1:T+\tau}$ given the history of the time series of person $v$:

$$\max_{\Phi, g} \sum_{v \in V} \sum_{T \leq T_v-\tau}\log p(x^{(v)}_{T+1:T+\tau} | x^{(v)}_{1:T} ;  g(v), \Phi).$$

The $\Phi$ refers to the set of learnable parameters of the model, which are shared between patients and learned jointly; $g(v)$ is the embedding of person $v$. However, the model still theoretically needs to take as much historical data as possible into account, which is both unrealistic and computationally infeasible. Intuitively, BG levels from weeks ago shouldn't influence today's BG levels. We formulate this assumption by only allowing the BG levels to depend on the most recent $t_0$ observations, or, mathematically, 
\begin{equation}
 (x^{(v)}_{T+1:T+\tau} |  x^{(v)}_{T-t_0+1:T}) \quad \independent \quad x^{(v)}_{1:T-t_0},
\end{equation}

where $\independent$ symbolizes independence and $(A|B)$ refers to conditioning random variable $A$ respect to the $B$. Then, define a summary vector $z_T^{(v)} \in \mathcal{R}^D$:
\begin{equation}
    z_T^{(v)}=f(x_{T-t_0+1}, x_{T-t_0+2}, \hdots, x_T, g(v)),
\end{equation}

such that:
\begin{equation}
    (x^{(v)}_{T+1:T+\tau} | z_T^{(v)}) \quad \independent \quad x^{(v)}.
\end{equation}

\begin{figure}
  \includegraphics[width=1\linewidth]{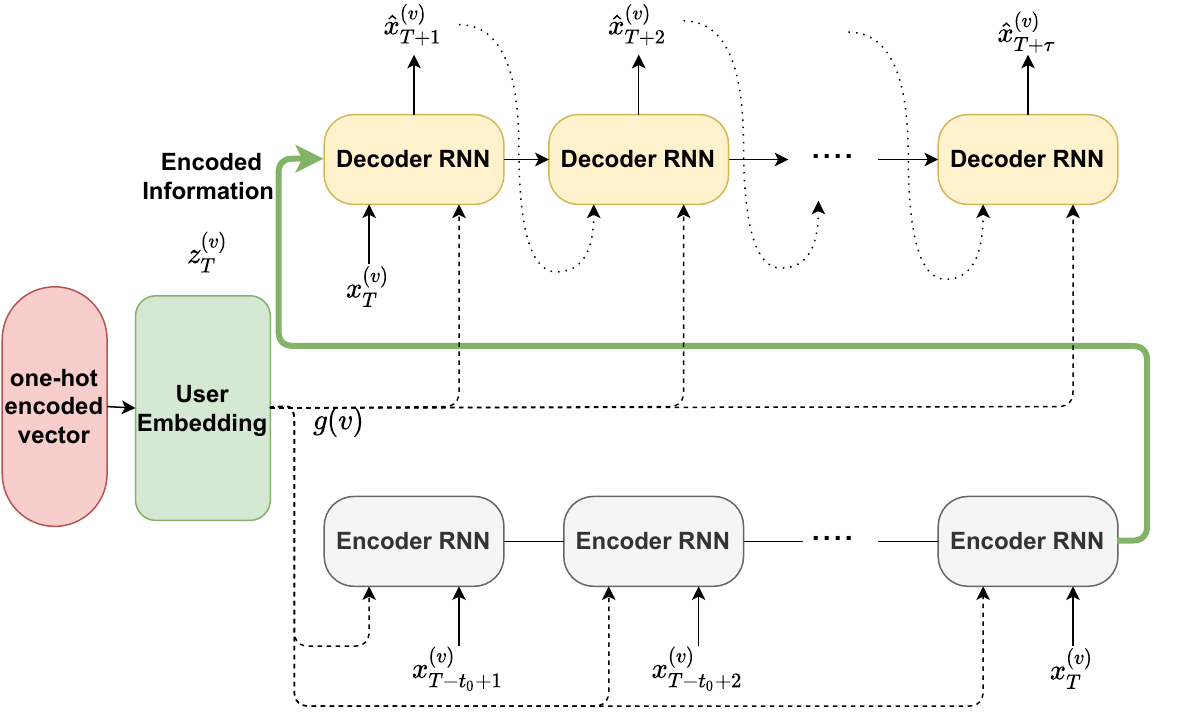}
  \caption{Deep Personalized Model}
  \label{DeepPe}
\end{figure}

The above assumption allows us to further simplify each term in the objective function:
\begin{align}
\log  p(x^{(v)}_{T+1:T+\tau} | & x^{(v)}_{1:T} ;  g(v), \Phi)\nonumber \\= 
& \log  p(x^{(v)}_{T+1:T+\tau} | x^{(v)}_{T-t_0+1:T} ;  g(v), \Phi) \nonumber\\=
& \log  p(x^{(v)}_{T+1:T+\tau} | z_T^{(v)} ;  g(v), \Phi). 
\end{align}
With the chain rule, we can further decompose the above objective term as:
\begin{equation} \label{decompose}
\begin{split}
\log  p(x^{(v)}_{T+1:T+\tau}  | z_T^{(v)}& ; g(v), \Phi)  \\
=\sum_{i=1}^{\tau} \log  p(x^{(v)}_{T+i} & | x^{(v)}_{T+1:T+i-1}, z_T^{(v)} ; g(v), \Phi).
\end{split}
\end{equation}

With this simplified objective, we define our deep architectures for learning both the summary vector $z_T^{(v)}$ and $\Phi$ for obtaining the log probability.

To learn $z_T^{(v)}$, we use a bidirectional RNN\cite{schuster1997bidirectional} (BiRNN) that sequentially reads the time series data of each individual as input. We employ BiRNN to capture both forward and backward trend information in the latent variable $z_T^{(v)}$. More specifically, for any $0 < j \leq t_0$, we set
\begin{equation}
h_j=\textit{BiRNN}([x_{T-t_0+j}^{(v)} ||g(v)],h_{j-1}). 
\end{equation}

\noindent The $||$ is the concatenation operation, and $h_i$ refers to the concatenation of hidden states from both the forward and backward paths. Then we let:
\begin{equation}
    z_{T}^{(v)}=\textit{tanh}(Wh_{t_0}), \quad \text{where} \quad W \in \mathcal{R}^{D\times N}, h_{t_0} \in \mathcal{R}^N
\end{equation}

and by $ \text{tanh} $ we refer to the hyperbolic tangent function. Lastly, we need to provide a parametric model for
\begin{equation}
     \log  p(x^{(v)}_{T+i}  | x^{(v)}_{T+1:T+i-1}, z_T^{(v)} ; g(v), \Phi).
\end{equation}

\noindent First, we approximate it by:
\begin{equation}
    \log  p(x^{(v)}_{T+i}  | \hat{x}^{(v)}_{T+1:T+i-1}, z_T^{(v)} ; g(v), \Phi),
\end{equation}

where $\hat{x}$ refers to our models approximation of the true $x$. We make another summary vector $s^{(v)}_{i-1}$ to capture the useful information from $\hat{x}^{(v)}_{T+1:T+i-2}$ for predicting $x^{(v)}_{T+i}$. In the following we explain the rest of the mathematical details:
\begin{align}
s^{(v)}_{i} & =\textit{RNN}([g(v)||\hat{x}^{(v)}_{T+i-1}], s^{(v)}_{i-1}), \\
s^{(v)}_{0} & = z_{T}^{(v)}, \quad \hat{x}^{(v)}_{T} = {x}^{(v)}_{T} \\
\hat{x}^{(v)}_{T+i} &=Q([s^{(v)}_{i}||g(v)]),
\end{align}
where the RNN just has the forward path and $Q$ is a fully connected neural network with no activation function at the last layer. Figure \ref{DeepPe} shows a summarized version of~our~model.

Our model can also be considered as a recurrent autoencoder, where $z_{T}^{(v)}$ is the encoded vector of the last $t_0$ time series observations. The decoder is made by putting a fully connected NN on top of the hidden states of the decoder RNN. The term $g(v)$ acts as an embedding vector for each patient, which is fed as input features to both the encoder and decoder RNN's. This lets the encoder and decoder behave differently for different patients and provides the opportunity for the RNN's in the model to show different behavior across different people.

\subsection{Capturing Long Range Dependencies}

In this section, we first investigate any possible time dependencies in the data and discuss possible explanations for these trends. Then, we propose a method to capture the patterns within the data with two key components: multi-head attention and added time features.

Following the statistical time-series literature, we use autocorrelation as a metric for time dependencies in the data. Autocorrelation, also known as serial correlation, is the correlation of a signal with its delayed version, defined~as:
\begin{equation}
    R(\tau)=\frac{E[(x_t - \mu)(x_{t+\tau} - \mu)]}{\sigma^2},
\end{equation}
where $x_t$ is the time series signals, $\mu$ and $\sigma^2$ are its mean and variance. In practice, we consider the empirical unbiased estimator (using the sample estimates of $\mu$ and $\sigma^2$) to calculate the autocorrelation.

\begin{figure}
  \includegraphics[width=1\linewidth]{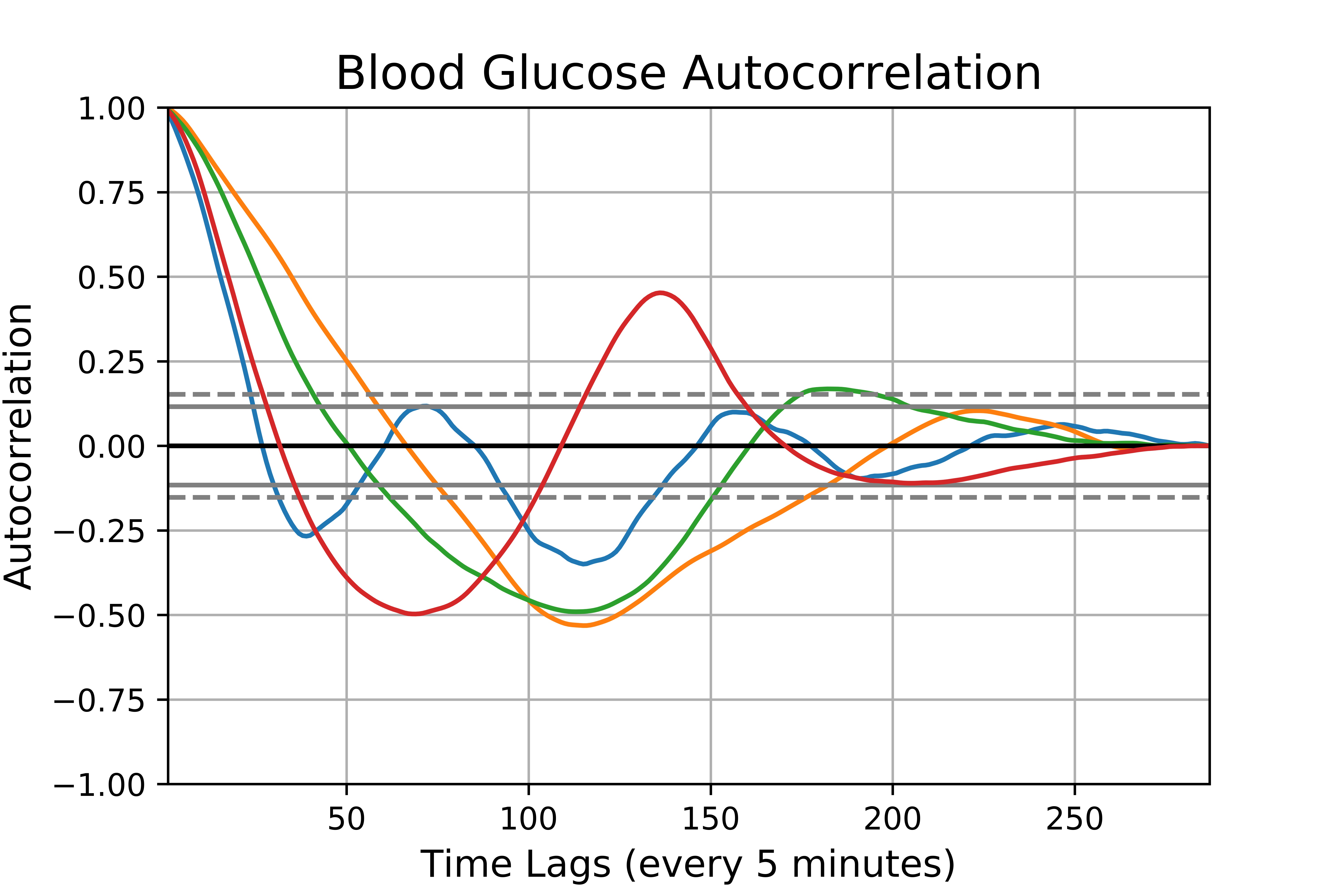}
  \caption{Autocorrelation graphs of randomly sampled time series over the course of one day. The dashed and solid horizontal lines correspond to the 99\% and 95\% confidence intervals for the correlation values around zero.}
  \label{autocor}
\end{figure}

A plot of the autocorrelation for a few randomly sampled time series is shown in Figure \ref{autocor}. The behavior is similar for these samples and those not shown; these are just representative of a common behavior. For the first few minutes (up to an hour and a half sometimes), there is significant positive correlation, as there is rarely a drastic change in BG in a short time interval. Therefore, our model needs to be able to capture this short-term trend. 

Later in the autocorrelation plot, some significant negative auto-correlation emerges. Capturing these much longer-term dependencies is a challenging task for standard models, like linear dynamical systems, hidden Markov models, and ARIMA. RNN's have been shown to perform better with long-term dependencies, but there is still the vanishing and exploding gradient problem \cite{pascanu2013difficulty}. Another issue is high variance in when the trend emerges, with significant negative auto-correlation appearing after a lag as large as 200 (16 hours). This irregularity in the pattern of negative correlation discouraged us from using convolutional modeling with fixed dilation \cite{oord2016wavenet} because convolutional models look for significant signals in fixed observation windows. Our solution captures this variability in negative trend by adaptively paying attention to different parts of the~time~series. 

\subsubsection{Multi-head Attention Layer}

We start by describing a single-head attention layer. The input to this layer is the set of all hidden states of the encoder $H=[h_1, h_2, \hdots , h_{t_0}]$ and the decoder hidden state at the previous time $s^{(v)}_{i-1}$. The output is a weighted sum of encoder hidden states across the whole encoding time. We mathematically describe the attention mechanism below:
\begin{equation}
e_{ij}=b(h_j, s^{(v)}_{i-1}),
\end{equation}
This $e_{ij}$ indicates the importance of the $j$'th hidden state of the encoder to the $i$'th output of the decoder. To make coefficients easily comparable across different hidden states, we normalize them using the softmax function across all choices of j:
\begin{equation}
\alpha_{ij}=\textit{softmax}_j(e_{ij})=\frac{\textit{exp}(e_{ij})}{\sum_s \textit{exp(}e_{is})},
\end{equation}
and we model the attention mechanism $a$ as following:
\begin{align}\label{att_equ}
e_{ij}=b(h_j, s^{(v)}_{i-1})=\textit{tanh}( r^T W[h_j||s^{(v)}_{i-1}]),   
\end{align}

\noindent where $r$ is a vector that needs to be learned and $r^T$ represents the transpose of $r$. Once obtained, the normalized attention coefficients are used as the weights for combining the hidden states. To obtain the final output of the attention layer, we apply an activation function $\sigma$ as:
\begin{align}
a_i=\sigma(\sum_{j}\alpha_{ij}h_j).
\end{align}
\begin{figure}
  \includegraphics[width=1\linewidth]{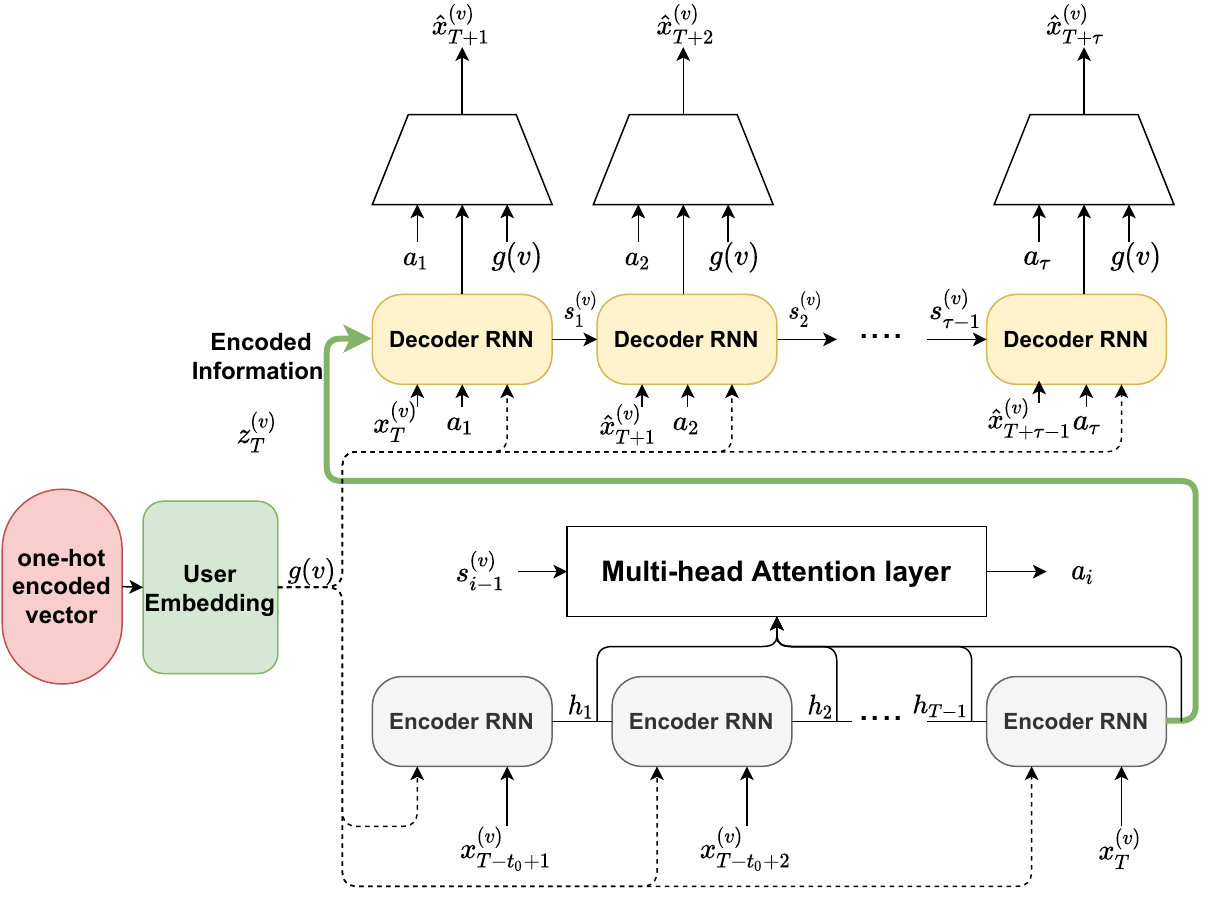}
  \caption{Deep personalized multi-head attention model}
  \label{atttime}
\end{figure}
In our experiments, we chose $ \textit{tanh} $ as our activation function. To extend the model to a multi-head attention model, we learn a different $r^{(k)}, W^{(k)}$ for each attention head. To be more precise, for the $k$'th head, there are separate $r^{(k)}, W^{(k)}$'s used to get $e_{ij}^{(k)}$ and $\alpha_{ij}^{(k)}$. Then, we let the output be
%To make multi-head attentions model we just need in equation \ref{att_equ}, for the $k$'th head, try $r^{(k)}, W^{(k)}$'s, to get consequently $e_{ij}^{(k)}$ and $\alpha_{ij}^{(k)}$, then we let:
\begin{align}
a_i=\sigma(\frac{1}{K} \sum_{k}\sum_{j}\alpha^{(k)}_{ij}h_j).
\end{align}
Then the attention vector $a_i$ is fed as an input feature to the decoder RNN at step $i$, as well as the fully connected NN. Analytically, we~let:
\begin{align}
s^{(v)}_{i} & =\textit{RNN}^{att}([a_i||g(v)||\hat{x}^{(v)}_{T+i-1}], s^{(v)}_{i-1}),\\
s^{(v)}_{0} & = z_{T}^{(v)}, \\
\hat{x}^{(v)}_{T+i} &=Q^{\textit{att}}([a_i||s^{(v)}_{i}||g(v)||\hat{x}^{(v)}_{T+i-1}]),
\end{align}
where the $ \textit{RNN}^{\textit{att}}$ and $Q^{\textit{att}}$ are again a forward path recurrent unit and a fully connected NN. We illustrate our final model architecture in the figure \ref{atttime}. 

The encoder in this new model does not have the burden of summarizing all the useful information in just a single vector $z^{(v)}_{T}$. The decoder can decide which parts of the input time series to pay attention to, instead of working with just a summarized vector that comes from the encoder. Lastly, in the new model, the gradient can pass through skip connections between encoder and decoder with the attention layer, which improves training and diminishes the exploding and vanishing gradient problem. 

\subsubsection{Time Features}

We explain why \textbf{time} is an important variable in the BG prediction task and how we can leverage it in our model. BG level is highly influenced by activities, such as eating and exercising, so a model that can better estimate eating time and/or exercising routine can provide more accurate forecasting. People tend to follow certain activity patterns based on environment; time is the simplest proxy for these patterns without more information.

Based on our observations from the data, we consider three time features. First, the hour of the day is important to capture scheduled daily patterns, like eating times and workout routines. Second, the day of the week is similarly considered as schedules sometimes vary slightly by day but remain fairly constant from week to week. Third, an indicator of weekend or not weekend is a crucial added emphasis on the day indicator, as schedules are often much less predictable during weekends (e.g. when there is no work or when travelling). We simply append these three time features to the input of each recurrent unit in our final model.

\subsection{Robust Training}

%In this section, we first describe the different training algorithms that we tried, then we explain some challenges in the training and propose our algorithm that can potentially avoid the problems. 

It's well known that training recurrent neural networks is a challenging task, and there have been several attempts to solve those problems, including \textit{teacher forcing} \cite{williams1989learning} and gradient clipping \cite{pascanu2013difficulty}. %Teacher forcing is a strategy for training RNN's that uses model output from a prior time step as an input. 
Standard RNN's use the model output from a prior time step as an input for predicting the next one when forecasting multiple steps into the future. Teacher forcing works by using the actual or expected output from the training dataset at the current time step as an input in the next time step instead. In training of our model, we try teacher forcing with an adaptive probability; however, we find the model becomes more fragile and gets worse results than when not using teacher forcing.

The next method we examine to aid training is gradient clipping. This forces the gradient values (element-wise) to have a maximum magnitude, "clipping" it to be that maximum if it is exceeded. Applying clipping to stabilize the batch gradients helps our RNN learn better. We further improve training by adaptively decaying the clipping threshold during training, as gradients should ideally shrink as training occurs and the model approaches an optimum.

Even with adaptive gradient clipping, we still observe large variance in the parameters' gradients for different mini-batch samples. Consequently, the model parameters fluctuate dramatically and are numerically unstable. By looking closer at the loss term for each time series sample in a mini-batch, we noticed the distribution of loss terms usually has some extreme outliers. The loss term outliers can overwhelm the total loss (i.e. the average) for the whole mini-batch. Hence, training based on a whole mini-batch is in fact just training on some "overwhelming" samples, so the model cannot properly learn the true general trends.

To circumvent this problem, we only consider the lower $\beta$-quantile of the loss terms related to the different samples within each mini-batch. Then, after removing the upper $1-\beta$ quantile of loss terms, we take the average as the loss of the whole mini-batch. In this way, we prevent the whole objective from being controlled by a few outliers. An added benefit is that it should ignore some of the unpredictable behavior (e.g. unexpected intake of a sugary food) during gradient calculation, making training more robust to unpredictable breaks in pattern. Empirically, we emphasize the improvement from this procedure in the experiments.

Some other methods, like the current state of the art model from \cite{fox2018deep}, consider the BG level as a discrete variable, so their model must also output a discrete variable. They use the cross-entropy loss for training to accommodate the discrete variables. This strategy also potentially avoids the outlier loss terms; however, the model does not have the right direction for the gradient for training the model when the distance between the estimated value and true value~is~large.

\section{Experiments}
In this section, we illustrate the performance of our model on real CGM data of 38 diabetes patients. To evaluate the performance of the model at the forecasting task, we need to decide on the length of the prediction window. Most of the existing works only considered 30 minute forecasts \cite{plis2014machine,turksoy2013hypoglycemia,zecchin2015jump,fox2018deep}, but we also consider one hour forecasts. This gives the patients more time to take proactive action in preventing hyper/hypo-glycemia by administering insulin or eating, which have delayed effects. However, as some methods were developed for the thirty minute intervals, we show that our method improves in the 15, 30, 45, and 60 minute forecasts.

Although our main goal is to provide forecasting trajectories of BG levels, it is important to evaluate the model performance when the patient experiences hypoglycemia ($<$ 70 mg/dl) or hyperglycemia ($>$ 180 mg/dl) separately. In our experiments, we include this by reporting the results for four different scenarios based on the BG value at forecasting time: Full (no constraint on BG level), Events (either  hypoglycemia or hyperglycemia), Hypo (hypoglycemia), and Hyper (hyperglycemia). 

We use the median of absolute percentage error (APE) and root mean square error (RMSE) over the forecasting window as two error metrics for the predictions. For $i$'th time step into future, we define APE as the median of $|x^{(v)}_{T+i} - \hat{x}^{(v)}_{T+i}|/x^{(v)}_{T+i}$, and RMSE as the square root of $\sum_{v} (x^{(v)}_{T+i} - \hat{x^{(v)}}_{T+i})^2 / n$, where $x^{(v)}_{T+i}$ denotes the true value and $n$ denotes the number of time series. Following \cite{fox2018deep} and unlike most of the previous work, we don't just calculate the error based on a single point in the future; instead, we use the average error based on all points in the forecast (e.g. six points for the 30 minute window). We prefer this method of evaluation as it is valuable to let a patient know about the trend as well. For example, if a patient's BG level rises toward the hyperglycemia range with an accelerating rate, the patient needs to be more alert to taking proactive action than the case when the rate~is~slowing.

\subsection{Data Description} 

To be consistent with previous work, we use the raw CGM data provided by \cite{fox2018deep}. Some of the CGM observations show drastic fluctuations of more than 40 mg/dl, which is physiologically unrealistic, so we removed them and considered them as errors. Unlike \cite{fox2018deep} we have not done polynomial interpolation for missing values for the large gaps. The data after cleaning consists of 399,302 observations (through time) for 38 patients. We temporally split the data to three sections: training, validation and testing. In fact, for each patient, we divided the data to these three sections with no overlap and a ratio of 20:1:1 with training first, validation second, and the testing data were the most recent observations. We use the validation set for early stopping and hyper-parameter tuning, but we do not use it for training the model.  
%For each person, we used the first 20/22 points for training, the next 1/22 points for validation, and the latest observations for the test set.

%We split the observations for each person into three sets with given ratios: 20 for training, 1 for validation, and 1 for testing. The earliest 20/22 observations are for training, the next for validation, and the most recent points for testing.

%
\begin{table*}[t]

\begin{adjustbox}{width=\textwidth, center}
  \begin{tabular}{l|cccc|cccc}
    \toprule 
    APE / RMSE & \multicolumn{4}{c|}{15 Minutes} & \multicolumn{4}{c}{30 Minutes} \\
    \cmidrule(r){2-5}  \cmidrule(r){6-9} 
    Algorithms &            Full &           Event &            Hypo &           Hyper &          Full &         Event &          Hypo &         Hyper \\
    \midrule
    ARIMA   &  3.31 / 5.98 &  3.27 / 7.54 &  7.83 / 6.13 &  3.13 / 7.68 &  5.89 / 11.21 &  5.43 / 11.87 &   12.14 / 9.41 &  4.98 / 12.83 \\
    RF: Rec    &  3.43 / 6.17 &  3.07 / 6.58 &  5.91 / 4.57 &  3.02 / 7.55 &  5.55 / 10.54 &  5.22 / 11.47 &   11.31 / 7.73 &  4.75 / 12.37 \\
    RF: MO     &  3.97 / 7.12 &  3.35 / 7.62 &  \textbf{5.73} / \textbf{4.31} &   3.14 / 7.90 &  6.13 / 11.46 &  5.23 / 11.56 &   12.09 / 9.23 &   4.90 / 12.12 \\
    PolySeqMo  &  3.07 / 5.55 &  2.96 / 6.72 &  7.27 / 6.32 &   \textbf{2.48} / 7.30 &    4.90 / 8.97 &  4.84 / 10.42 &  13.38 / 11.97 &  \textbf{4.03} / \textbf{10.32} \\
    % DeepState & \textbf{2.84} / \textbf{5.08} & \textbf{2.87} / \textbf{5.91} & 5.89 / 4.61 & 2.51 / \textbf{6.49} & 4.52 / 9.08 & 4.71 / 10.33 & 10.32 / 8.42 & 4.04 / 10.94 \\
    Our Method (MSE)              &   3.02 / 5.40 &  3.03 / 6.64 &  8.07 / 6.23 &   2.59 / 6.65 &   4.79 / 9.18 &  4.82 / 10.36 &   11.89 / 9.99 &  4.08 / 11.19 \\
    Our Method (Robust) &   \textbf{2.90} / \textbf{5.26} &   \textbf{2.90} / \textbf{6.32} &  6.36 / 4.92 &  2.57 / 6.62 &   \textbf{4.47} / \textbf{8.91} &   \textbf{4.55} / \textbf{10.30} &     \textbf{9.96} / \textbf{7.50} &  4.04 / 10.89 \\
    \bottomrule
   
    \end{tabular}%

    \end{adjustbox}
\caption{The median forecasting errors, in terms of APE and RMSE, for 15 and 30 minute prediction windows} \label{tab:1}
\end{table*}
\begin{table*}[t]

\begin{adjustbox}{width=\textwidth, center}
  \begin{tabular}{l|cccc|cccc}
    \toprule 
    APE / RMSE & \multicolumn{4}{c|}{45 Minutes} & \multicolumn{4}{c}{60 Minutes} \\
    \cmidrule(r){2-5}  \cmidrule(r){6-9} 
    Algorithms &            Full &           Event &            Hypo &           Hyper &          Full &         Event &          Hypo &         Hyper \\
    \midrule
    ARIMA    &   7.84 / 14.63 &  6.93 / 15.89 &  16.32 / 12.91 &  6.84 / 18.16 &   9.85 / 17.65 &  8.91 / 19.86 &  19.94 / 14.53 &  8.51 / 22.17 \\
    RF: Rec    &   7.11 / 14.10 &  6.89 / 15.96 &  15.08 / 10.41 &  6.77 / 18.46 &   9.04 / 17.15 &  8.97 / 20.36 &  18.84 / \textbf{12.43} &  8.68 / 23.41 \\
    RF: MO     &  8.06 / 14.85 &  6.91 / 15.75 &  17.09 / 13.58 &   6.39 / 16.90 &  10.22 / 18.27 &   8.61 / 19.90 &  21.64 / 17.36 &  7.99 / 21.58 \\
    PolySeqMo  &  6.83 / 12.32 &  6.46 / 14.52 &   18.51 / 17.30 &   \textbf{5.42} / \textbf{14.20} &   8.55 / 15.68 &  8.27 / 18.81 &  22.86 / 21.87 &   \textbf{6.77} / \textbf{18.30} \\
    % DeepState & 6.52 / 12.79 & 6.43 / 14.37 & 15.71 / 13.64 & 5.67 / 15.88 & 8.35 / 15.43 & 8.14 / 18.29 & 16.91 / 13.12 & 6.80 / 18.59 \\
    Our Method (MSE)        &   6.44 / 12.60 &  6.65 / 15.25 &  15.69 / 13.09 &  5.63 / 15.44 &   8.17 / 15.67 &  8.29 / 19.37 &  18.72 / 16.26 &  6.99 / 19.22\\
    Our Method (Robust) &  \textbf{6.27} / \textbf{11.74} &  \textbf{6.21} / \textbf{14.09} &   \textbf{13.08} / \textbf{9.81} &  5.45 / 15.04 &   \textbf{7.92} / \textbf{14.81} &   \textbf{8.01} / \textbf{17.80} &  \textbf{15.97} / 12.57 &  6.82 / 19.07 \\
    \bottomrule
   
    \end{tabular}%

    \end{adjustbox}
\caption{The median forecasting errors, in terms of APE and RMSE, for 45 and 60 minute prediction windows} \label{tab:2}
\end{table*}

\subsection{Compared Algorithms}

We compared our method to both classical statistical methods and deep learning based approaches, including the state of the art method. We did not compare with models that were already shown to be inferior for BG prediction. During training, we allow each algorithm to utilize up to the previous 190 data points (approximately~16~hours).

\begin{itemize}
\item \textbf{ARIMA:} A traditional statistical method for time series prediction is auto-regressive integrated moving average (ARIMA), and they have been previously used in forecasting BG level \cite{plis2014machine}. As the name suggests, there are three parts to the model: an auto-regressive part, an integration, or differencing, part, and a moving average part. The auto-regressive (AR) part is simply regressing the current value on the past $p$ values of the time series. The integrated part is designed to account for non-stationary trends. Lastly, the moving average piece models is a regression of the current value on the past $q$ error terms. Combining these correctly can be quite powerful for~consistent trends.

\item \textbf{Random Forests:} Random forests are a standard baseline ensemble method for prediction. It builds many prediction trees and then averages the predictions from each tree to get a final ensembled prediction. They also have been used for BG forecasting \cite{sudharsan2014hypoglycemia}. There are two possible implementations for predicting multiple time points into the future. One way is for the RF to output the whole prediction window time series as a vector (RF: MO for multi-output). An alternative to attempt to incorporate time in the prediction is to recursively do one-step ahead prediction with a sliding window (so that the input dimension remains consistent) and the output is one dimensional (RF: Rec).

\item \textbf{PolySeqMo:} The current state of the art model \cite{fox2018deep} uses a recurrent auto-encoder model and has shown superior results to the standard statistical methods. PolySeqMo uses an encoder to summarize the history of BG levels; it uses the encoded vector to learn the coefficients of a polynomial with the decoder. The method then uses the coefficients to provide forecasting based on the fitted polynomial. The model does not provide personalized forecasting but instead learns a generic model for the entire population. Since they got the best result when they fixed the degree of the polynomial to one (i.e. linear forecasting), we follow the same setup for comparison. They fix the prediction window to 6 time lags (30 minutes); however, we changed the prediction window to 12 (one hour) for a longer comparison as well.

\end{itemize}
\begin{figure*}[htb]
%\begin{adjustbox}{width=\textwidth, center}
    \centering % <-- added
\begin{subfigure}{0.25\textwidth}
  \includegraphics[width=\linewidth]{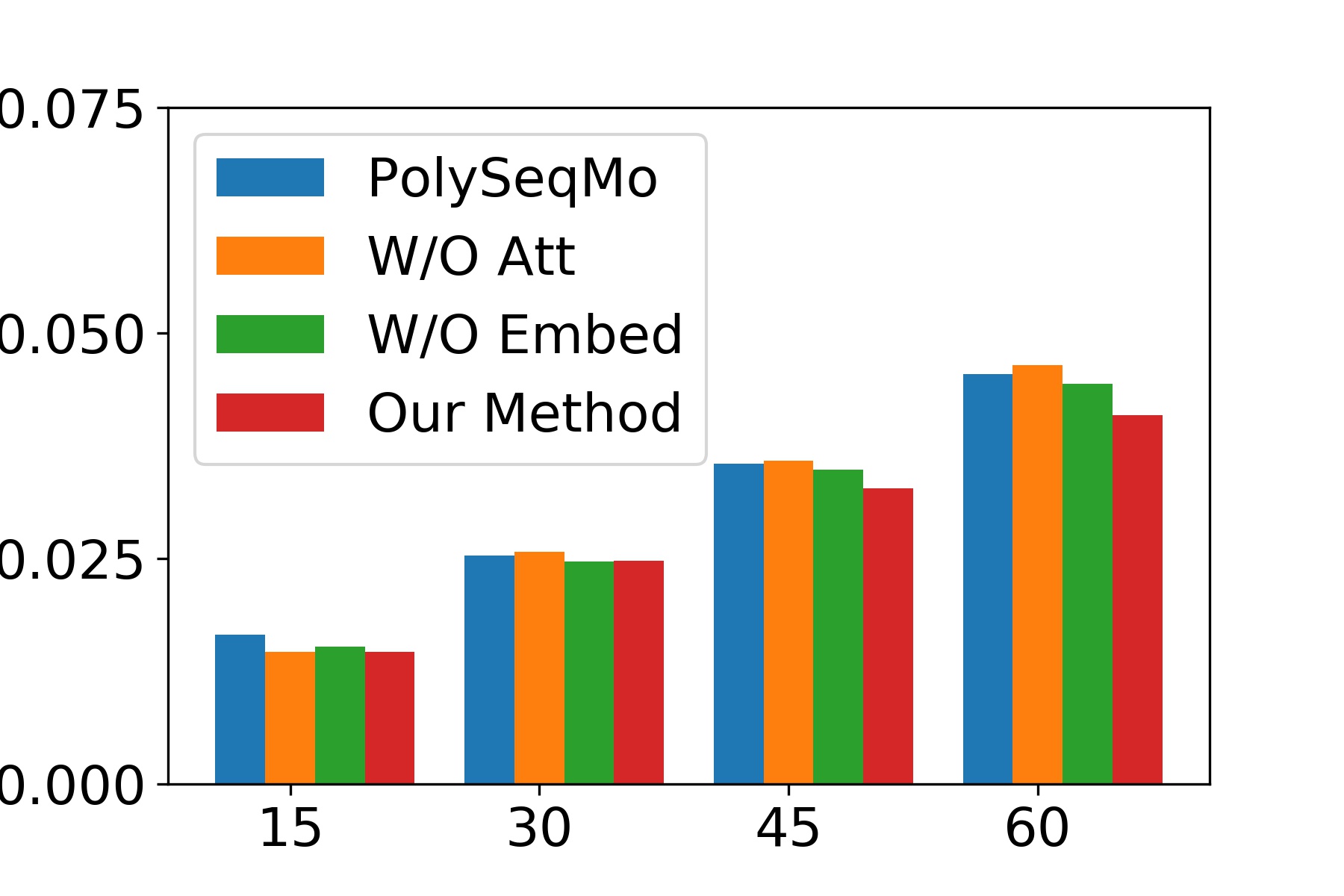}
  \caption{The first quartile of APE}
  \label{fig:1}
\end{subfigure}\hfil % <-- added
\begin{subfigure}{0.25\textwidth}
  \includegraphics[width=\linewidth]{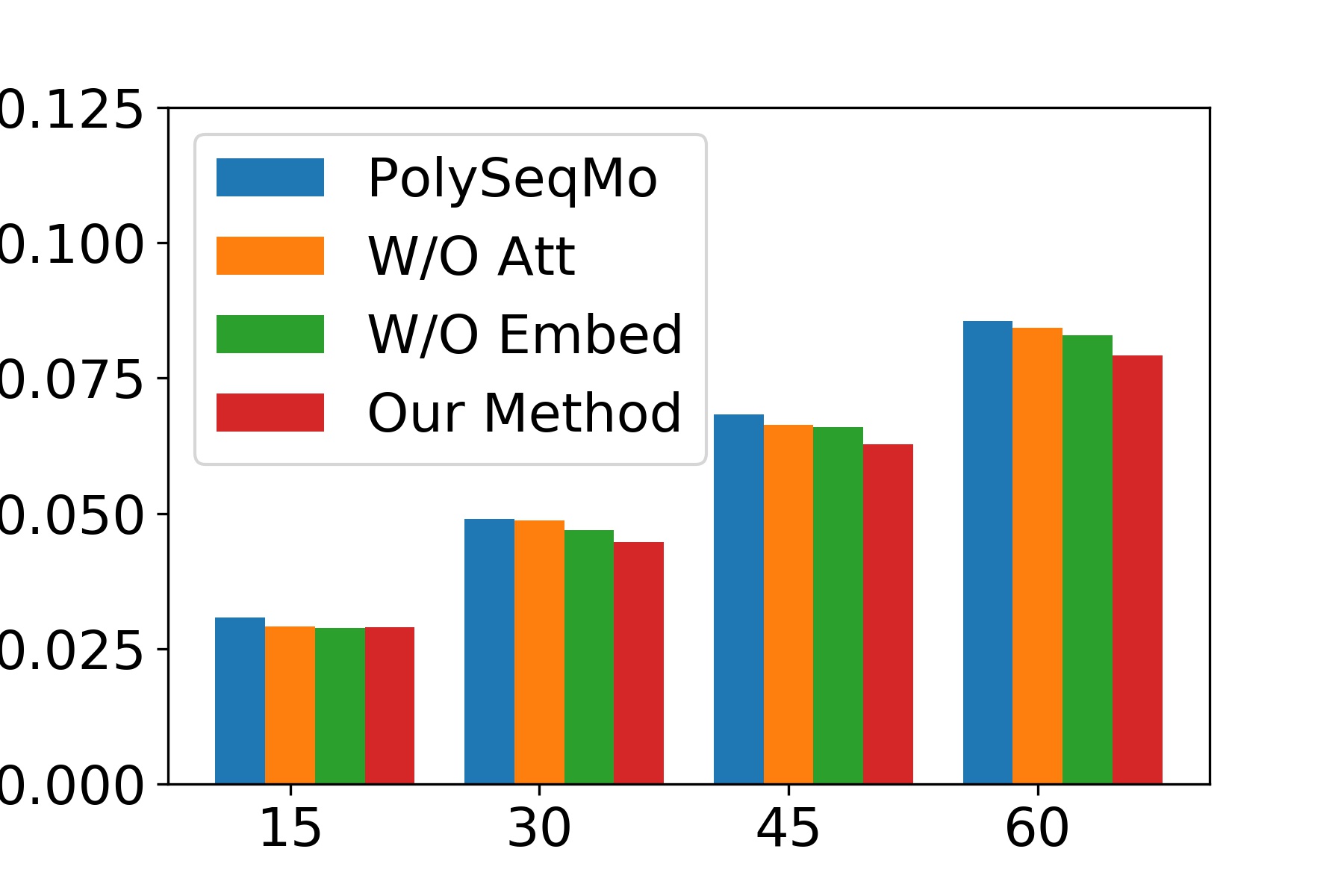}
  \caption{The median of APE}
  \label{fig:2}
\end{subfigure}\hfil % <-- added
\begin{subfigure}{0.25\textwidth}
  \includegraphics[width=\linewidth]{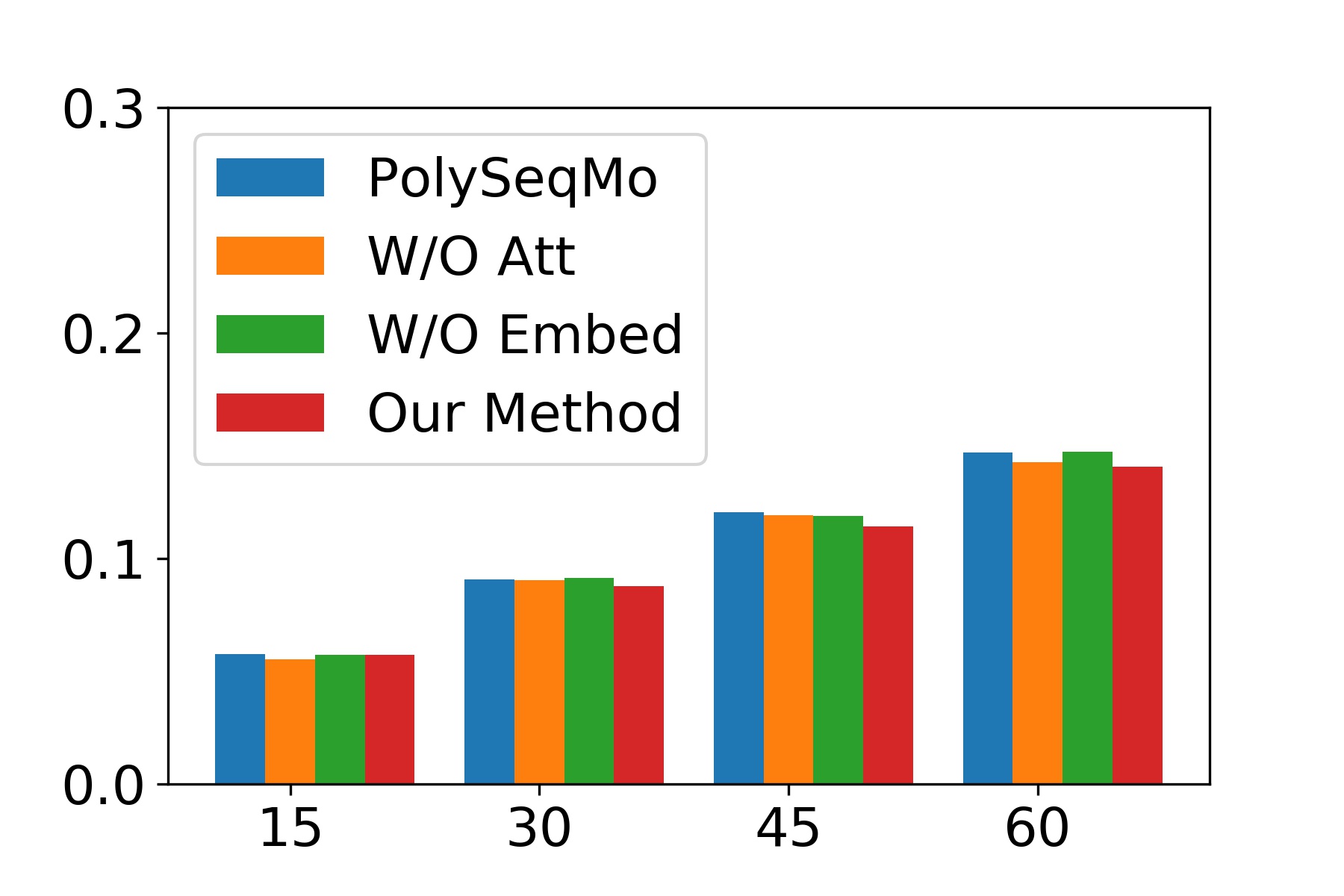}
  \caption{The third quartile of APE}
  \label{fig:3}
\end{subfigure}

\medskip
\begin{subfigure}{0.25\textwidth}
  \includegraphics[width=\linewidth]{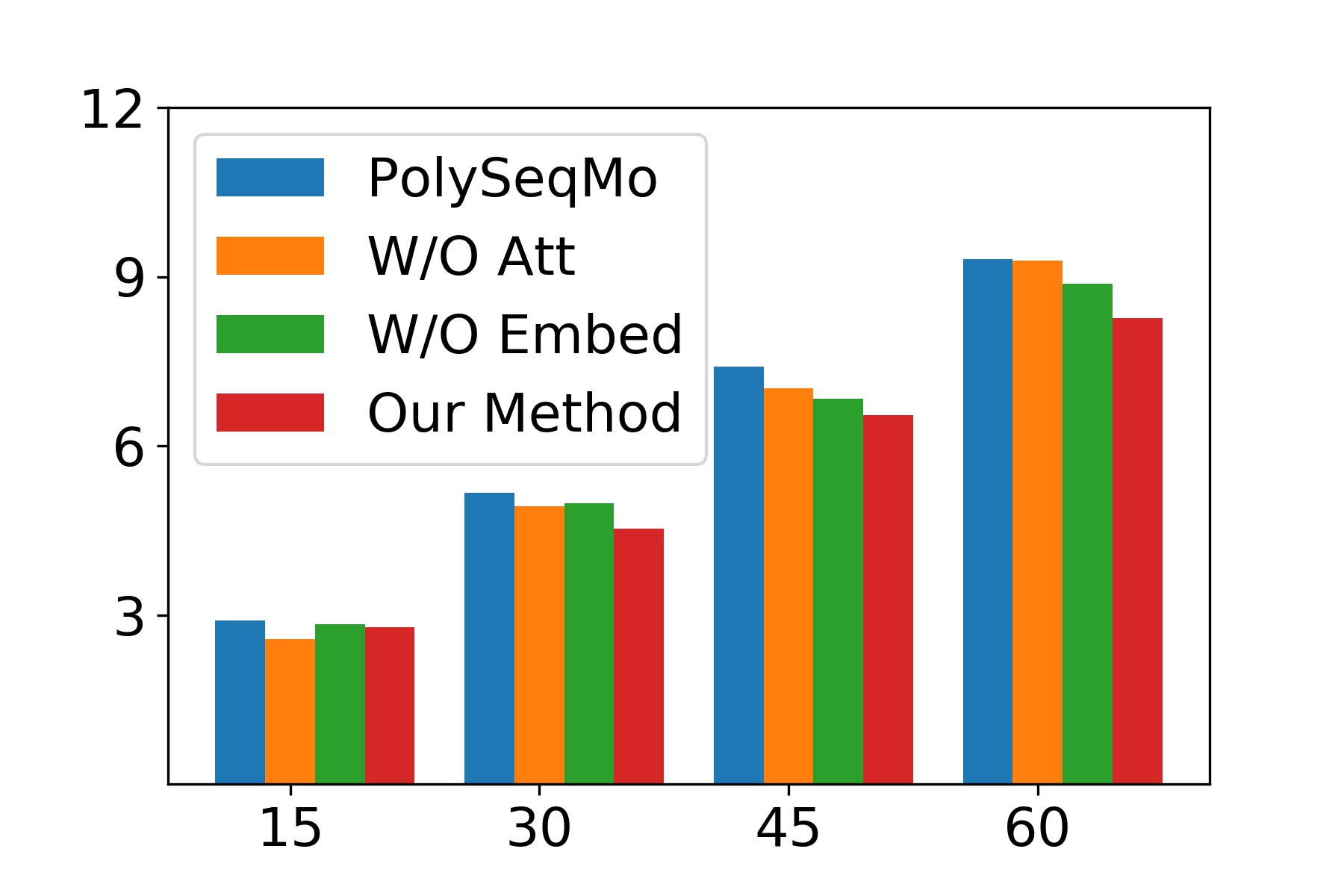}
  \caption{The first quartile of RMSE}
  \label{fig:4}
\end{subfigure}\hfil % <-- added
\begin{subfigure}{0.25\textwidth}
  \includegraphics[width=\linewidth]{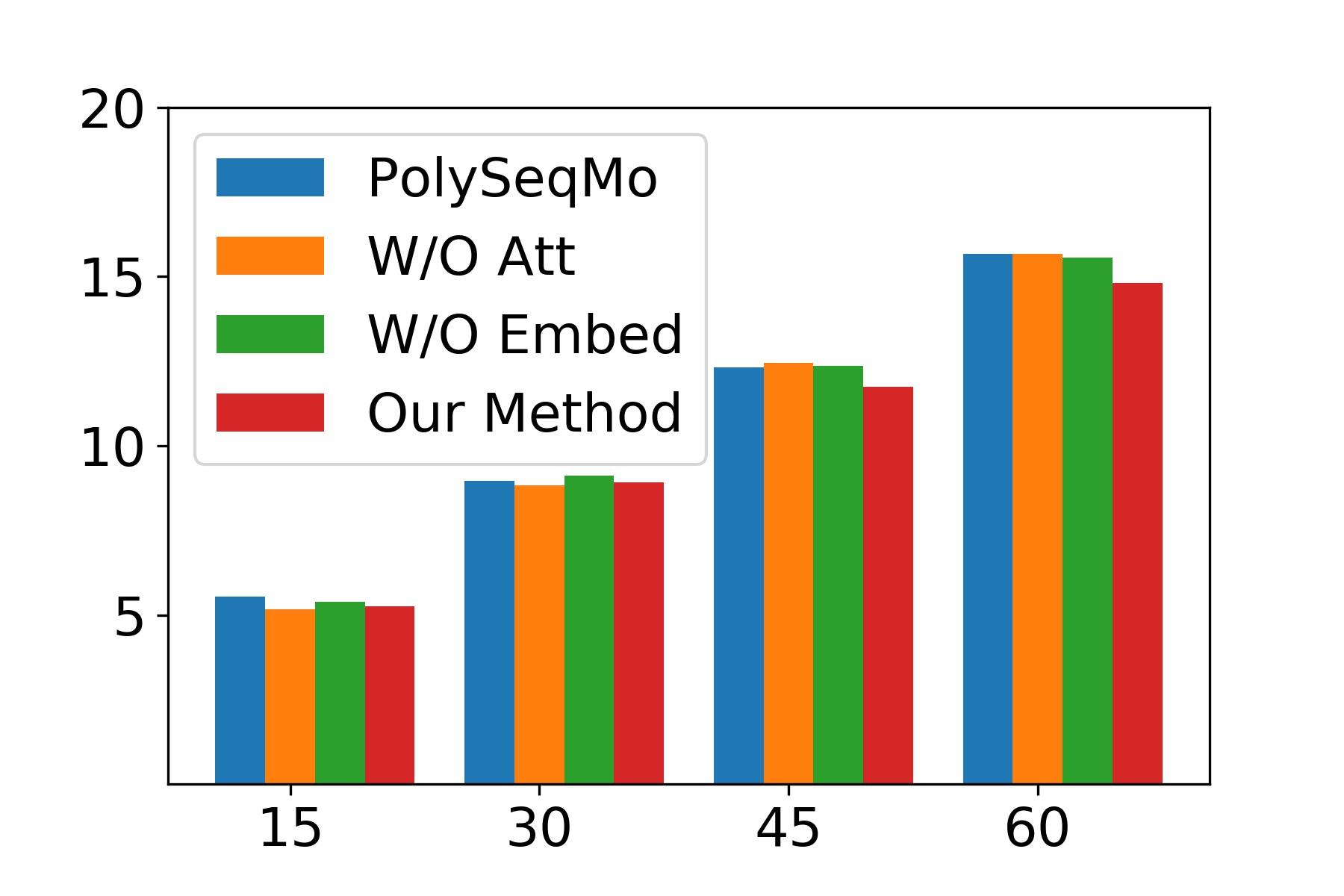}
  \caption{The median of RMSE}
  \label{fig:5}
\end{subfigure}\hfil % <-- added
\begin{subfigure}{0.25\textwidth}
  \includegraphics[width=\linewidth]{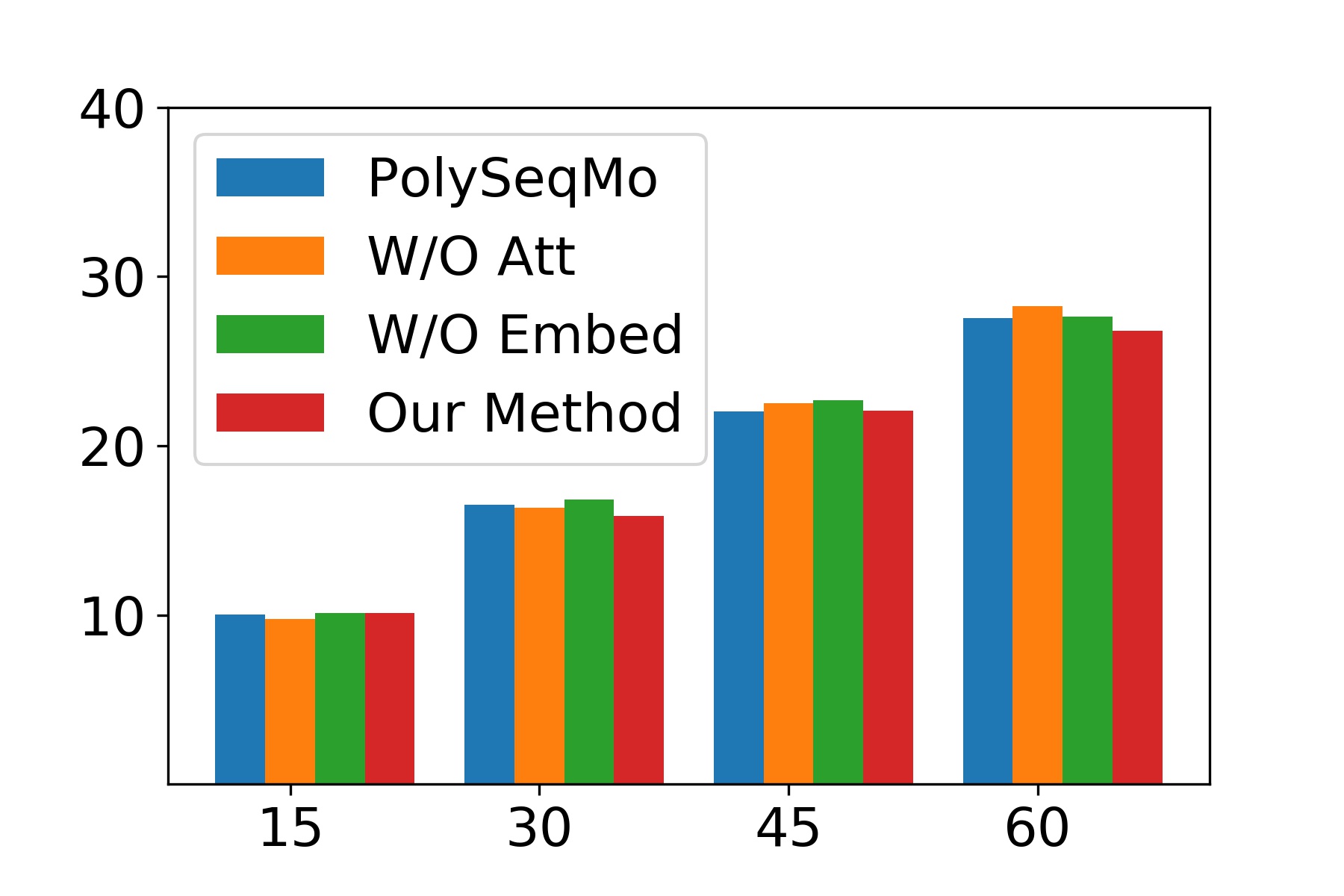}
  \caption{The third quartile of RMSE}
  \label{fig:6}
\end{subfigure}
\caption{Ablation study forecasting results in terms of APE and RMSE for 15, 30, 45, and 60 minutes into the future}
\label{fig:images}
%\end{adjustbox}
\end{figure*}

\subsection{Implementation Details}

We implement our model in Pytorch and use RAdam \cite{liu2019variance} for the optimization with the default parameters. For our method, we do adaptive gradient clipping with an initial value of 2 and a decay rate of 0.99 for each epoch, and we do early stopping based on the validation set. For the encoder, we use a bidirectional Gated Recurrent Unit (GRU) \cite{chung2014empirical} with 120 hidden dimensions and for the decoder we use a feed forward GRU with 30 hidden dimensions. A single layer fully connected neural network with 60 hidden units transforms the decoder output into the explicit forecast values. Our embedding consists of applying a linear transformation to the one-hot patient encoding vector to obtain a 5 dimensional embedding. We initialize each parameter in our model with a normal distribution with mean 0 and standard deviation 0.1. The~top 10 percent of loss terms are not used while training the model~($\beta = 0.9$). 

To be able to capture long-term dependencies, we set the encoder length to be 190 in our experiment and for the forecasting window, we use 12 time steps (one hour) into the future. We followed the choice in \cite{fox2018deep} to use a 10-step input size for the RF baselines, which are implemented using ScikitLearn \cite{pedregosa2011scikit}. The code, the trained models, and the data will be (is) available at the Github page of the authors\footnote{\url{https://github.com/rezaarmand/glucose_att_rnn}}.

\subsection{Results}

Tables \ref{tab:1} and \ref{tab:2} summarize the forecasting errors of the different methods for 15, 30, 45, and 60 minutes into the future. The variance of the reported medians in the tables is less than 0.01 since we  had a large (over 3,000) test set. As is shown in the "Full" column, our method performs best overall for predicting BG levels in the future. Our method also performs the best in term of predicting the hypoglycemia APE for more than 15 minutes into the future. Short term health complications of hypoglycemia can be deadly, so having an early warning is crucial. Hypoglycemia forecasting is the most difficult to predict in terms of APE (as it has the largest values), but our method improves on it by about 15\% for 30 minutes or longer into the future. In the cases that our method is not better than the others, the difference is often insignificant. 

To highlight the importance of robust training, we compare our full robust method to our method without robust training (Our Method (MSE)). Tables \ref{tab:1} and \ref{tab:2} clearly show that adding robust training improves performance in all cases. Even without the robust training, our method performs better than PolySeqMo in many scenarios. It is also of note that due to the categorical training of PolySeqMo, our robust training addition is not directly applicable.

PolySeqMo has another disadvantage besides the overall performance: their best results come from only fitting a linear forecast. As we argued in the introduction, knowing the trend is quite important to understand the urgency of required action. Our method can capture general trends (not limited to polynomial trends) due to the direct use of added inputs and both GRU and NN layers for decoding.
\subsection{Ablation Study}

Finally, we provide an ablation study to stress the importance of using both an attention mechanism and personalized embeddings. We add two additional versions of our method: one without attention mechanism ('W/O Att') and the other by excluding embedding vectors ('W/O Embed'). For comparison, we plot these alongside both our full model and the previously described PolySeqMo model in Figure \ref{fig:images}. The result is based on using the entire test data set and correspond to the "Full" columns in Tables \ref{tab:1}, \ref{tab:2}. The plots provide quartile estimates to show the variability of errors and clearly affirm that all parts of our model are necessary to better capture the long-term dependencies in blood glucose levels.

\section{Conclusion}
In this paper, we study the problem of BG trajectory forecasting without any need of exogenous information, such as carbohydrate intake or exercise. Our algorithm has three main components that we motivate by analyzing real data. One component is using the idea of embedding vectors to provide personalized prediction. By having shared parameters across the patients, the model can alleviate the need for a huge set of observations for each patient. Another component is designed to capture long term dependencies within the time series. To achieve this better than vanilla recurrent auto-encoders, we add an attention mechanism and provide time features as inputs. Our method reserves the ability to take other information as inputs to make more informative blood glucose forecasting. The last component is a more robust training algorithm to further improve our results. 

Compared to a state of the art method, PolySeqMo (which has already shown effectiveness over other widely used methods), our approach demonstrated a consistent improvement for forecasting window of all sizes up to one hour, indicated by smaller error metrics on a real world large dataset consisting of 400K continuous blood glucose measurements. Such results highly suggest the efficacy of our proposed approach in forecasting blood glucose levels and more generally in providing better care for people living with diabetes. 

For future work, we would investigate the probabilistic forecasting solutions to the problem. The idea is to not only provide a point estimate of the future values but also a confidence bound about how certain the model is about its estimations. We plan to do that by changing the objective to a log-normal loss function and learning the variance of the estimated forecast with a recurrent autoencoder or using Bayesian neural network.

\section{Acknowledgements}
The authors thank Ran Duan for many useful discussions and guidance. The authors thank the anonymous reviewers and area chair for their valuable comments and suggestions that have helped us to improve the paper.
\bibliographystyle{IEEEtran}
\bibliography{sample-base}

\end{document}